\pdfoutput=1

\documentclass[11pt]{article}

\usepackage[]{acl}

\usepackage{times}
\usepackage{latexsym}
\usepackage{graphicx}
\usepackage{booktabs}
\usepackage{multirow}
\usepackage[T1]{fontenc}

\usepackage[utf8]{inputenc}

\usepackage{microtype}
\usepackage{amsmath}
\usepackage{mathtools}
\usepackage{tcolorbox}
\usepackage{enumitem}

\definecolor{GoodEdit}{RGB}{68, 108, 174}
\definecolor{BadEdit}{RGB}{245, 106, 108}
\newtcbox{\Goodbox}{on line,colframe=white,colback=GoodEdit!20!white, boxrule=0.5pt,arc=0pt,boxsep=0pt,left=2pt,right=2pt,top=1pt,bottom=1pt,coltext=blue!45!black}

\newtcbox{\Goodboxhigher}{on line,colframe=white,colback=GoodEdit!20!white, boxrule=0.5pt,arc=0pt,boxsep=0pt,left=2pt,right=2pt,top=1pt,bottom=0.2pt,coltext=blue!45!black}
\newtcbox{\Badboxhigher}{on line,colframe=white,colback=BadEdit!20!white, boxrule=0.5pt,arc=0pt,boxsep=0pt,left=2pt,right=2pt,top=1pt,bottom=0.2pt,coltext=red!45!black}
\newtcbox{\Highlightboxhigher}{on line,colframe=white,colback=GoodEdit!20!white, boxrule=0.5pt,arc=0pt,boxsep=0pt,left=2pt,right=2pt,top=1pt,bottom=0.2pt,coltext=green!45!black}
\newtcbox{\Highlightbox}{on line,colframe=white,colback=GoodEdit!20!white, boxrule=0.5pt,arc=0pt,boxsep=0pt,left=2pt,right=2pt,top=1pt,bottom=1pt,coltext=green!45!black}
\newtcbox{\Badbox}{on line,colframe=white,colback=BadEdit!20!white, boxrule=0.5pt,arc=0pt,boxsep=0pt,left=2pt,right=2pt,top=1pt,bottom=1pt,coltext=red!45!black}

\title{Modeling Intensification for Sign Language Generation: A Computational Approach}

\author{
    Mert İnan\textsuperscript{\rm 1}\thanks{~ The first three authors have equal contribution.} ,
    Yang Zhong\textsuperscript{\rm 1}\textsuperscript{*},
    Sabit Hassan\textsuperscript{\rm 1}\textsuperscript{*},
    Lorna Quandt\textsuperscript{\rm 2},
    Malihe Alikhani\textsuperscript{\rm 1} \\
    \textsuperscript{\rm 1} Computer Science Department, School of Computing and Information, \\
    University of Pittsburgh, Pittsburgh, USA\\
    \textsuperscript{\rm 2} Educational Neuroscience Program, Gallaudet University, Washington, D.C, USA\\
    \texttt{ \{mert.inan, yaz118, sah259, malihe\}@pitt.edu,} \\ \texttt{ lorna.quandt@gallaudet.edu}
}

\usepackage{todonotes} 

\begin{document}
\maketitle

\begin{abstract}

End-to-end sign language generation models do not accurately represent the prosody in sign language. A lack of temporal and spatial variations leads to poor-quality generated presentations that confuse human interpreters. In this paper, we aim to improve the prosody in generated sign languages by modeling \textit{intensification} in a data-driven manner. We present different strategies grounded in linguistics of sign language that inform how intensity modifiers can be represented in gloss annotations.
To employ our strategies, we first annotate a subset of the benchmark PHOENIX-14T, a German Sign Language dataset, with different levels of intensification. We then use a supervised intensity tagger to extend the annotated dataset and obtain labels for the remaining portion of it. This enhanced dataset is then used to train state-of-the-art transformer models for sign language generation. We find that our efforts in intensification modeling yield better results when evaluated with automatic metrics. Human evaluation also indicates a higher preference of the videos generated using our model.

\end{abstract}

\begin{figure}[t!]
\small
    \centering
    \begin{tabular}{cc}
         \includegraphics[height=3.6cm]{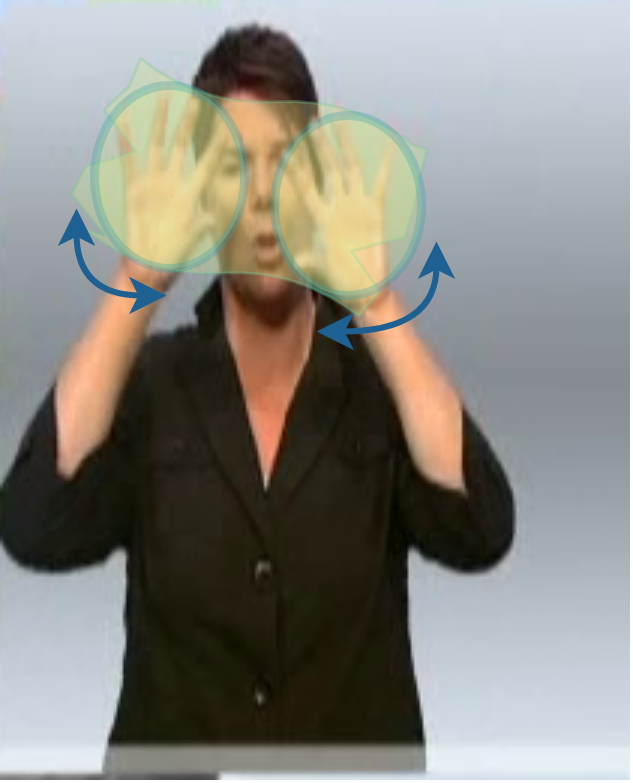} & \includegraphics[height=3.6cm]{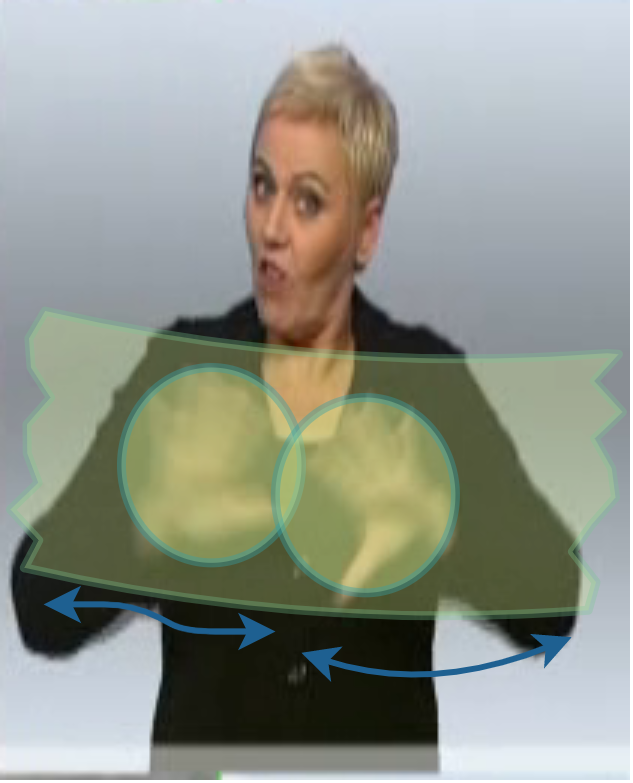}\\
         less clouds & very cloudy \\
         \textbf{\texttt{WOLKE}} & \textbf{\texttt{WOLKE}} \\
         \toprule
         \midrule
         10 video frames & 17 video frames\\[0.1ex]
          \midrule
         Sign Not Repeated & Sign Repeated \\[0.1ex]
          \midrule
         No Delay & Delayed Beginning\\[0.1ex]
          \midrule
         Smaller Space Use & Larger Space Use\\[0.1ex]
         \bottomrule
    \end{tabular}
    \caption{In sign language, modifiers are represented spatially and temporally. Here, two signers from PHOENIX-14T manually sign German "less clouds", and "very cloudy". Both of these signs have the same gloss representation: \texttt{WOLKE} (cloud in German).
    They are figuratively the same sign, but the duration, repetition, temporal pauses, and continuations determine the exact meaning. This information is lost during sign language translation and evaluation.}
    \label{fig:main_fig}
\end{figure}

\section{Introduction}

Similar to spoken languages, sign language has rich grammar rules and unique linguistic structures \cite{yin-etal-2021-signed,emmorey2001sign}. Elements of prosody such as rhythm, stress, or lengthening play important roles in distinguishing meaning and signaling intensification in sign language (Figure~\ref{fig:main_fig}), similar to spoken languages \cite{Brentari2018Production}. Thus, it is important for sign language generation (SLG) systems to be able to learn accurately from the data and generate presentations that respect prosody. 

Much of the current study on prosodic markers such as intensifiers \cite{ Bolinger+2013, Rett2008DegreeMI, davidse_2011} are based on linguistic theories of spoken languages and need to be adapted to signed languages, as prosody is represented in the visual modality \cite{wennerstrom_2001}. Semantic differences are signaled in the visual modality using spatial and temporal presentations such as iconicity, gesture duration, as well as temporal pauses \cite{degree-mod}. Such distinctive properties present challenges in SLG systems to generate presentations with better prosody.

Several SLG systems have been proposed in recent years motivated by their importance to the Deaf and Hard of Hearing (DHH) communities \cite{surrey848809, zelinka2020synthesis, stoll2020text2sign, Saunders_2021_ICCV}. Transformer-based models \cite{saunders2020progressive} have been shown to outperform other neural models \cite{stoll2020text2sign} in generating sign language from gloss annotations \textemdash a shortened approximation of spoken language that has mappings to signs. 
One of the key limitations of the state-of-the-art models is that the prosody of the sign videos generated by these models does not change with the semantics of the signs \cite{duarte2021how2sign}. 
%
In this paper, we take a step toward the goal of modeling prosody in sign language generation by modeling \textit{intensification}. We refer to intensification as the presence of \textit{intensity modifiers} that quantify nouns, adjectives or adverbs in a sentence. The intensity modifiers can either be an amplifier (e.g., lot of rain) or a diminisher (e.g., little rain). Studies in the linguistics of signed languages show that intensity modifiers change the duration and tactile emphasis in the produced sign \cite{degree-mod}. Thus, intensification modeling can impact prosody of generated signs. However, this potential of intensification is not realized within current models because they depend on gloss representation. Intensity modifiers are often excluded in gloss representation because they are a sparse approximation of spoken language.  As shown in Figure \ref{fig:main_fig}, the spatial and temporal properties of signs differ dramatically even when they map to the same gloss. State-of-the-art models cannot be aware of this temporal and spatial manipulation by modifiers if they are not represented in the gloss training data.

Our initial analysis of the PHOENIX-14T \cite{camgoz2018neural}, a German Sign Language dataset, reveals that 23\% of the data has at least one adjective or adverb in the text transcript, but none in the gloss representation. Since adjectives and adverbs (e.g., little) often act as intensity modifiers, they are likely to be under-represented in the gloss as well. This observation motivates the need for explicit modeling of intensification in the gloss representation and modifying state-of-the-art models to incorporate this additional information. We hypothesize this to have an overall improvement in the models’ performance both quantitatively in terms of automated metrics and qualitatively in terms of human evaluation. To this end, drawing on linguistics and cognitive science studies of sign language, we
\vspace{-\topsep}

\begin{enumerate}
\setlength{\itemsep}{0pt}

  \setlength{\itemsep}{1pt plus 1pt}
    \item introduce gloss enhancement strategies grounded in linguistics that respect the role of modifiers with various levels of intensity.
    \item present a supervised tagging model to improve a given gloss dataset with modifier intensity levels using strategies we have identified.
    \item make available an enhanced version of the PHOENIX-14T dataset where the glosses are tagged with intensity levels of modifiers.
    \item incorporate modifier information into the Progressive Transformer (PT) model. We also propose a novel model that can dynamically select the generated poses with different gloss enhancement as input. We make our code and data publicly available.\footnote{\url{https://github.com/Merterm/Modeling-Intensification-for-SLG}}
\end{enumerate}

\section{Related Work}

\paragraph{Prosody of Sign Language}
Prosodic information in sign language has been studied through the lenses of cognitive sciences and linguistics. Using brain images, Newman et al., \shortcite{NEWMAN2010669} show that prosodic signed information is processed by signers in much the same way as it is by hearing speakers. In \cite{doi:10.1177/00238309990420020101}, the intertwined nature of prosody is observed in a multifaceted manner for semantics, neurological basis and syntactic understanding of sign languages. Nicodemus et al., \shortcite{nicodemus_2009} note that prosodic markers play an important role as delimiting units during the production and perception of the signs. These works study the importance of prosodic markers during the producing and processing sign language by humans from a cognitive science perspective. In our work, 
we model intensification as a prosodic marker computationally.

In linguistics research, studies have focused on the relationship between prosody and syntax in sign language \cite{Sandler2010ProsodyAS}, role of prosody in identifying breakpoints in discourse, and detection of salient events \cite{Ormel2012ProsodicCO}. Sandler et al. \shortcite{Sandler2020SignLP} suggest that pragmatic notions related to information structure are a part of prosody in sign language.
Although there has been limited work that highlight the importance of intensity modifiers in sign language prosody \cite{degree-mod}, our work is the first data-driven empirical study that studies a large dataset, annotates, then quantifies and characterizes data-driven strategies for modeling intensification. 
Our work is the first that presents a Transformer-based model for intensification as a step toward modeling prosody.

\paragraph{Sign Language Generation}
Many works have looked at sign language processing, such as coreference resolution \cite{yin-etal-2021-including} or gloss augmentation for translating gloss into text \cite{moryossef-etal-2021-data}. However, prosody is still understudied in the field of sign language generation and processing. 

The primary aim of SLG is generating sign poses from texts. Earlier work has explored methods to generate animated avatars \cite{cox2002tessa, Glauert2006VANESSAA,McDonald2015AnAT} from speech or text inputs, but were restricted by the rule-based systems and the modest size of sign pose libraries. More recently, with the introduction of larger corpora such as PHOENIX-14T \cite{camgoz2018neural}, and advanced deep learning model architectures, generating more accurate and expressive human skeletal sequences from spoken language transcripts or annotated glosses has become possible
\cite{surrey848809, stoll2020text2sign, zelinka2020synthesis, Saunders2020AdversarialTF, saunders2020progressive, Saunders_2021_ICCV} while also including facial expressions \cite{viegas2022including}. Yet, none of these works attempt at modeling intensification or any other indicator of prosody in hand gestures. Our work is the first that combines linguistic and cognitive findings and proposes a deep learning model that dynamically selects intensification strategies to generate skeletons with variations for different levels of intensifiers based on augmented glosses.
\section{Intensification in Sign Language}
\label{sec:heuristics}


Gloss annotations in the German Sign Language weather forecast corpus, PHOENIX-14T, are simple German words that often do not capture the subtleties of sign language. For example, "very cloudy" and "slightly cloudy" are both represented by a single gloss "WOLKE" (CLOUD). Our analysis shows that in 23 percent of the data, the gloss representation does not contain any adjectives or adverbs present in the text transcript. Since intensity modifiers are usually adjectives/adverbs that quantify intensity of other words, we expect them to be missing from the gloss representation as well. Hence, in order for the model to represent intensity modifiers in its latent space, it is necessary to include them in the training data.
\subsection{Gloss Enhancement Strategies}
Focusing on what our data presents, we analyze the best ways of representing intensity modifiers in gloss annotations based on linguistic theories, cognitive science and neuroscience perspectives of intensities in sign language. We discover that the choice of order for the additional gloss modifier tokens matter. Linguistic analysis of American Sign Language also shows the importance of this. 

Wilbur et al. \shortcite{degree-mod} explain that depending on the degree of the adjective, there is a "sharp movement to a stop" in the final timing of the sign, which is coined as \textit{end-marking}. They also show that the initial time interval of a sign also gets modified with a slight pause in the beginning and a faster continuation of the sign, which is termed as a \textit{delayed-release}. Also, there exists other datasets with different annotation schemes, one of which --Public DGS Corpus-- uses a gloss annotation convention where the phonemes and synonyms that have different signs contain a number that is added as a suffix to the end of the gloss \cite{konrad_reiner_2020_1860}. Finally, as described by \cite{10.3316/informit.209984769714521} during the end-marking and elongation phase, a sign might be reiterated to mark the intensification.

\begin{figure*}[!t]
    \centering
    \includegraphics[scale=0.8]{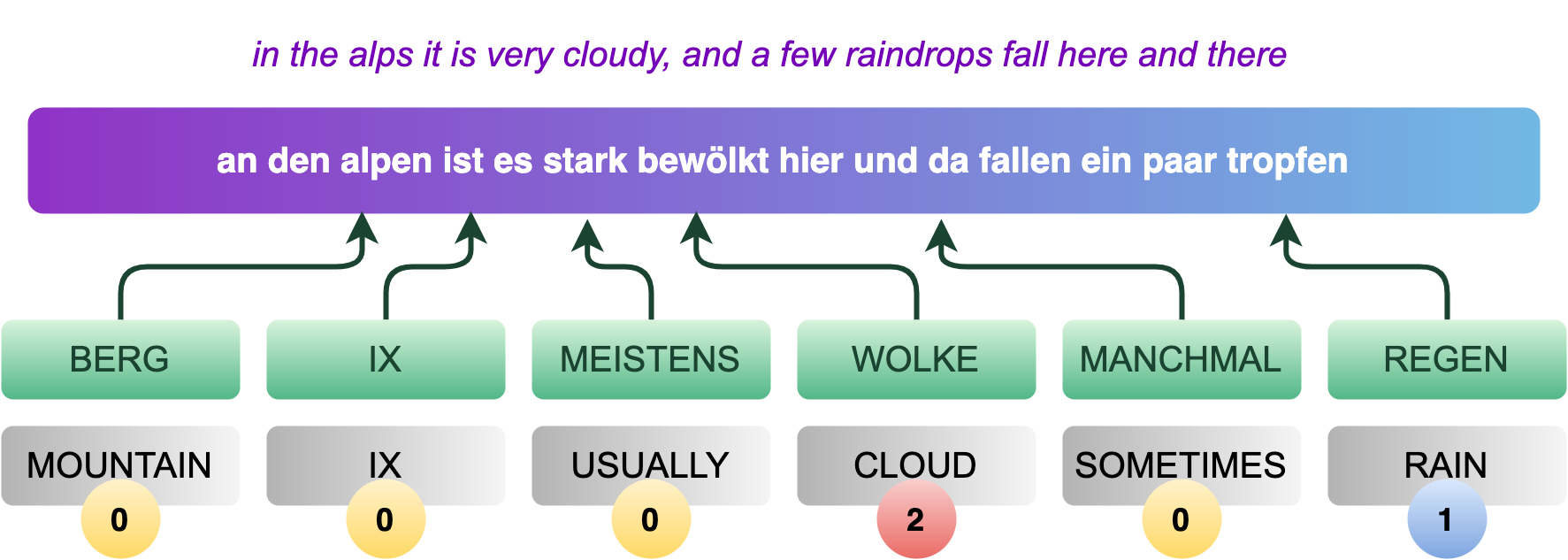}
    \caption{This figure shows an example annotation. German transcript text and gloss are provided as context along with their English translations. Each English gloss in the sentence are tagged with 0, 1, 2, corresponding to the degree of intensification.} 
    \label{fig:annotation}
\end{figure*}

Inspired by these previous works in linguistics of signed languages and in analyzing the dataset with sign language researchers, we came up with four strategies to better represent intensity modifiers in glosses. We use these strategies in four alternative ways, as shown in Table \ref{tab:augment_example} and are introduced below:
\begin{itemize}\setlength{\itemsep}{0pt}

    \item \textbf{End-Marking}, where an additional token of \texttt{<HIGH-INT>} or \texttt{<LOW-INT>} is added \textit{after} the intensity-modified gloss to represent the change in the final timing of the sign as shown in \cite{degree-mod}.
    \item \textbf{Delayed Release}, where the additional intensity modifier token of \texttt{<HIGH-INT>} or \texttt{<LOW-INT>} is added \textit{before} the original gloss, as described in \cite{degree-mod} to represent the delayed release in the initial timing of the sign. 
    \item \textbf{Suffixation}, where an \texttt{INT} suffix is added at the end of the gloss with an additional numerical value (1 or 2) corresponding to the degree of intensification. 
    This is analogous to the Public DGS Corpus annotation \cite{konrad_reiner_2020_1860}.
    \item \textbf{Reiteration}, where we repeat the intensity-modified gloss token twice to capture this in the gloss representation as described by \cite{10.3316/informit.209984769714521}.
\end{itemize}

\begin{table}
\small
\centering
\begin{tabular}{l|l}
\toprule
\textbf{Approach} & \textbf{Example} \\
\midrule
Text & very cloudy \\
Original Gloss & \texttt{WOLKE} (cloud)  \\
\midrule
Suffi.&  \texttt{WOLKE-INT2} \\
End-mark. &  \texttt{WOLKE <INT2>} \\
Delay.-rel. &  \texttt{<INT2> WOLKE} \\
Suffix.-reiter.  & \texttt{WOLKE-INT2 WOLKE-INT2} \\
\bottomrule
\end{tabular}
\caption{Gloss Enhancement examples.}
\label{tab:augment_example}
\end{table}

\subsection{Data Annotation}

We start by selecting a subset of the publicly available PHOENIX-14T dataset \cite{camgoz2018neural} for the annotations of intensity modification. 


\paragraph{Data Sampling.} Initial analysis demonstrates that gloss annotations tend to ignore the adjectives/adverbs, which are signals of intensity modification. We hypothesize that for samples where the number of adjectives/adverbs is zero in gloss annotations but more than zero in texts, the intensity information is more likely to be missed. We use Spacy \cite{spacy2} part-of-speech (POS) tagger to tag the text and gloss pairs, then utilize the hypothesis mentioned above to filter the data. In the end, we acquire 1557 samples in the train set, 132 samples in the development set, and 157 samples in the test set. Afterwards, the gloss sequences are split into individual gloss tokens. These gloss tokens are paired with the full text transcripts, which yields a total of 12.8K gloss token to sentence pairs -- 10.8K from the 1557 instances in train, 1K from the 132 instances in dev and 1K from the 157 instances test set.

\paragraph{Annotation Protocol.}
For each of the gloss token to sentence pair,  we ask at least one annotator to assign labels to the gloss token from the following categories: (i) \texttt{2} as ``high intensity'' if there is an intensity modifier such as ``high'' in the text surrounding the gloss; (ii) \texttt{1} as ``low intensity'' if the intensifier in the text marks a low degree intensity; or (iii) \texttt{0} if there is no corresponding modifiers in the text transcripts.\footnote{We translated the German transcriptions and glosses into English using the Google Translate API \url{https://cloud.google.com/translate}} Figure \ref{fig:annotation} shows an example of the annotation. 

\paragraph{Annotator Agreement.}
Three expert annotators were recruited according to the rules and regulations of our institution's human-subject board. 
To assess the inter-annotator agreement,
we randomly sampled 700 token-sentence pairs and asked all three annotators to annotate. The resulting Fleiss' Kappa \cite{Fleiss1974StatisticalMF} coefficient is 69.2, which suggests a substantial agreement among the annotators. 

\begin{figure*}[ht]
    \centering
    \includegraphics[scale=0.8]{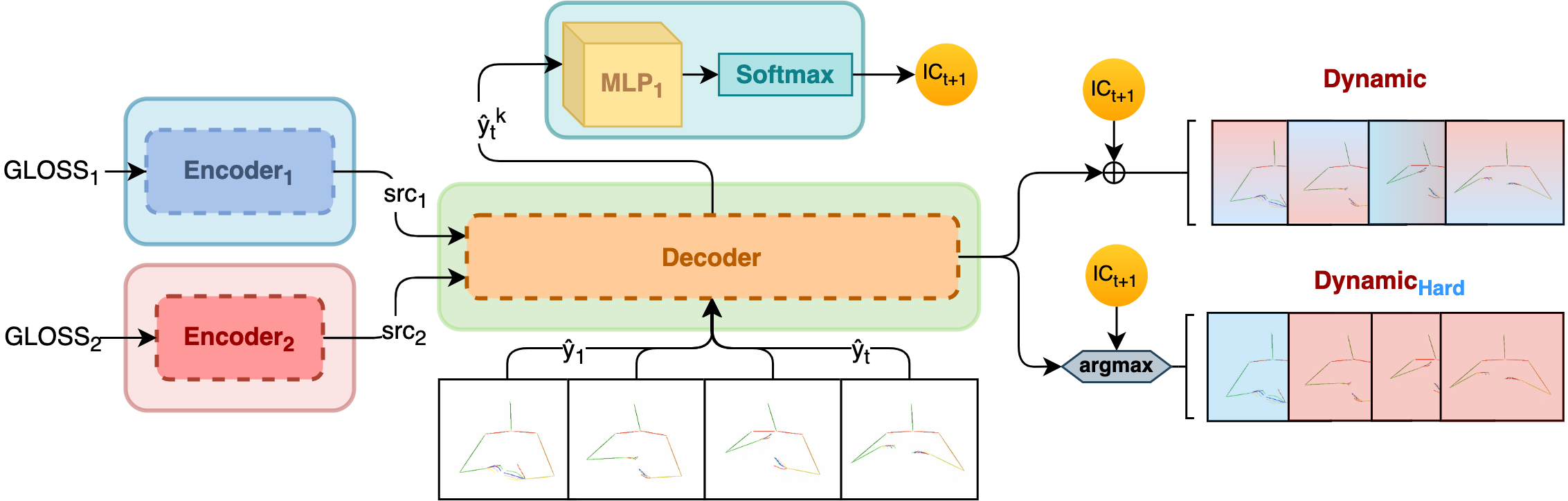}
    \caption{This figure shows the architecture of the Dynamic Selection model. The overall architecture is similar to the Progressive Transformer, except having two Encoders to select between two different types of strategies. MLP layer is the decisive step on selecting the strategy from the encoders. Dynamic model uses a weighted mixture of the decoder outputs (represented with a gradient of blue and red). Dynamic$_{hard}$ uses an argmax to pick a source.}
    \label{fig:dynamic_model}
\end{figure*}

\subsection{Full Corpus Intensity Enhancement}
\label{sec:corpus_construct}
Utilizing the annotated pairs, we train a battery of classifiers to automatically predict the gloss labels for the remaining data points. Having an automated classifier saves us resources that would otherwise be needed to tag the whole dataset. 

\paragraph{Classifiers.} We frame the task as a text pair classification problem. Given the original text transcript and a gloss token, the goal is to predict a label from: \texttt{0} (no intensity modification), \texttt{1} (low degree intensity) and \texttt{2} (high degree intensity).  We experimented with multiple classification baselines, including fastText \cite{joulin2017bag}, Bidirectional LSTM and two versions of fine-tuned BERT \cite{devlin-etal-2019-bert} models -- German BERT (G-BERT) and multilingual BERT (M-BERT).
All models are trained on the manually annotated 10.8K training pairs and results are reported on the 1K test subset.


\begin{table}[b]
\small
\begin{center}
\begin{tabular}{c|c|c|c|c}
\toprule
\bf{Model} & \bf{Features} & \bf{Prec.} & \bf{Recall} & \bf{F1}\\
\midrule
FastText & embed & 60.5 & 62.0 & 61.0\\
BiLSTM & embed & 62.1 & 66.6 & 64.1\\
G-BERT & -- & \bf{74.3} & 74.2 & 74.2\\
M-BERT &  -- &    74.2 & \bf{76.4} & \bf{75.3}\\
\bottomrule
\end{tabular}
\end{center}
\caption{GLOSS intensifier classification results. Embeddings for FastText and BiLSTM are learned during training.}
\label{tab:gloss-classification}
\end{table}

Table \ref{tab:gloss-classification} shows the experiments with different classifiers. 
Fine-tuned transformers G-BERT and M-BERT outperform others by a large margin.
The performance improvement of M-BERT compared to G-BERT is statistically significant according to a permutation test.

\paragraph{Error Analysis of Gloss Enhancement}
We manually categorize 100 errors made by our best classifier, M-BERT. The key observations are: i) 30\% of the errors are due to ambiguity that annotators may have for hard cases. E.g., "The wind blows weakly to moderately" can be annotated as either low-intensity (weakly) or no-intensity (moderately).
ii) aligning gloss tokens with text can be difficult (24\%). For example, in "partial snow or freezing rain", the classifier may consider "partial" to be aligned with rain, assigning it the label of "low-intensity" (should be "no-intensity"). Further, presence of negation (e.g., "not much rain") and multiple occurrences of the same word (e.g., "in the Bergland, it snows partly, on the alps it snows for a long time.") can make alignment a difficult task for the classifier, and
iii) 12\% of the errors can be attributed to noise in original PHOENIX-14T data. E.g., the gloss representation can contain tokens that are not related to the transcript. We could not assign a specific category to 34\% of the errors.

 \paragraph{Enhancement.} We tag all the remaining glosses with the best-performing classifier, M-BERT, in the original PHOENIX-14T dataset. We end up with four versions of enhanced gloss sequences by incorporating the aforementioned strategies in section \S \ref{sec:heuristics}, namely \textit{Suffixation}, \textit{End-marking}, \textit{Delayed Release} and \textit{Suffixation with Reiteration}. 

\section{Model}
In this section, we first introduce a baseline model that has been widely adopted for the sign language generation task (section \S \ref{PT-baseline}). To better model the signer's dynamic intensification choices during sign production, we further propose a dynamic selection model (Figure \ref{fig:dynamic_model}) that makes use of inputs with different intensity modification strategies.

\subsection{Progressive Transformer Baseline} \label{PT-baseline}
\label{sec:pt_baseline}
The main goal of the sign language generation model is to transform a gloss or text sequence into skeletal pose coordinates per each frame of the signing video. Formally, given a gloss sequence 
\begin{math}{X = [x_1, ... x_N]}\end{math}, a sign language generation model aims to learn the conditional probability \begin{math}p = ( Y | X) \end{math} where Y represents the corresponding skeletal pose coordinate sequence \begin{math}{Y = [y_1, ... y_T]}\end{math}. We use the
Progressive Transformer (PT) \cite{saunders2020progressive} model as our baseline.
The model employs an encoder-decoder architecture to generate a sign language sequence  \begin{math}{\hat{Y} = [\hat{y}_1, ... , \hat{y}_T]}\end{math} in an auto-regressive manner. The encoder is composed of L transformer layers, each with one Multi-Head Attention (MHA) and a feed-forward layer. The computed representation of the source sequence is fed into a modified transformer decoder,  which employs a counter-based decoding mechanism to guide the generation of continuous joint sequences $ \hat{y}_{1:T}$ and to decide the end of the generated sequence. This strategy can be formulated as:
\begin{equation}
    [\hat{y}_{t+1}, \hat{c}_{t+1}] = PT(\hat{y}_{t} | \hat{y}_{1:{t-1}}, x_{1:N} )
\end{equation} where $\hat{y}_{t+1}$ and $\hat{c}_{t+1}$ are the generated joint sequence and the counter value for the generated frame $t+1$. 
The model is trained using the mean square error (MSE) loss between the generated sequence $\hat{y}_{1:T}$ and the ground truth $y_{1:T}$: 
\vspace{-.2cm}
\begin{equation}
L_{MSE} = \frac{1}{T} \sum_{i=1}^{T}( y_{i} - \hat{y}_{i})^2
\end{equation} It is worth noting that, as stated by \cite{huang2021fast}, the proposed decoding mechanism provides weak supervisions with the initial ground-truth frame and guided counter sequences during the inference time.

\begin{table*}[h!]
\small
    \centering
    \setlength\tabcolsep{2.1pt}
    \renewcommand{\arraystretch}{1.1}
    \begin{tabular}{l|cccc|cccc|cccc}
    \toprule
        \multicolumn{12}{c}{\textbf{\textit{DEV SET}}} \\
        & \multicolumn{4}{c}{with intensification (248)} & \multicolumn{4}{|c}{without intensification (271)} &\multicolumn{4}{|c}{full} \\
        \midrule
         &   B\textsubscript{1} & B\textsubscript{4} & RG  & BS  &  B\textsubscript{1} & B\textsubscript{4} &  RG   & BS  &   B\textsubscript{1} &  B\textsubscript{4} & RG & BS \\

         \midrule
         
         Baseline & 25.07 & 6.24 & 22.61 & 72.20 & 35.46 & 17.98 &  36.84 & 77.46 & 29.92 & 11.90 & 30.05 & 74.95 \\
         \midrule
         Suffix. & 25.72 & 6.71 & 24.03** & 72.61 &  37.73** & \textbf{19.35}** & \textbf{38.92}** &  \textbf{77.88} & 31.32* & \textbf{12.81} & {31.81}** & 75.36 \\
         Delay.-rel.  & 27.03** & 6.67 & 24.31** & {72.97} & \textbf{37.75}** & 18.39 & 38.55** & 77.84 &  \textbf{32.03}** & 12.35 & 31.74**  & 75.51\\
        End-mark.  & \textbf{27.32}** & \textbf{7.29} & {24.46}** & {72.52}& 36.48 & 18.08 & 37.26 & 77.42 & 31.59* & 12.51 & 31.15 & 75.08 \\ 
        Suff.-reiter. & 26.23* & 6.74 & {24.78}** & 72.78 &  35.98 & 17.97 & 37.92 & 77.74 &  30.77 & 12.20 & 31.64* & 75.37\\
        \midrule
        Dynamic & 25.88 & 6.52 & 23.82*  & 72.54 & 35.65 & 17.80 & 37.59 & 77.86 & 30.44 & 11.99 & 31.01 & 75.32\\
        Dynamic\textsubscript{hard} & 26.01 & 6.36 & \textbf{24.98}**  & \textbf{73.06} & 36.35 & 18.25 & 38.75** & 77.87  & 30.83 & 12.20 & \textbf{32.17}** & \textbf{75.57} \\

        \toprule
        \multicolumn{12}{c}{\textbf{\textit{TEST SET}}} \\
        & \multicolumn{4}{|c}{with intensification (314)} & \multicolumn{4}{|c}{without intensification (328)} &\multicolumn{4}{|c}{full}\\
        \midrule
        &   B\textsubscript{1} & B\textsubscript{4} & RG  & BS  &  B\textsubscript{1} & B\textsubscript{4} &  RG   & BS  &   B\textsubscript{1} &  B\textsubscript{4} & RG & BS \\
      
        \midrule
        Baseline & 25.28 & 5.92 & 21.98 & 72.02 & 35.17 & 17.40 &  35.97 & 76.85 & 29.86 & 11.51 & 29.13 & 74.49 \\
        \midrule
        Suffix. & 26.31 & 6.54 & {24.56}** & 73.10 & 33.70 & 17.14 & 34.60 & 76.87 &  29.73 & 11.71 & 29.69 & 75.03 \\
        Delay.-rel. & 19.33 & 3.43 & 16.29 & 69.56 & \textbf{36.07} & 17.53 & 36.49 & 77.31 &27.08 & 10.27 & 26.61 & 73.52\\
        End-mark. & 23.98 & {6.67} & 22.38 & 72.09 & 34.94 & 17.28 &  35.27 & 76.60 & 29.05 & 11.73 & 28.96 & 74.39 \\
        Suff.-reiter. & 25.04 & 6.24 & 23.41* & \textbf{73.13} & 34.85 & \textbf{17.63} & {36.43} & \textbf{77.65} &  29.58 & {11.74} & {30.06} & \textbf{75.44}\\
        \midrule
        Dynamic & 26.06 & {6.79} & 23.89** & 72.76 & 35.42 &  17.21 & \textbf{36.53} & 77.42 & \textbf{30.39} & {11.79} & \textbf{30.34} & 75.13 \\
        Dynamic\textsubscript{hard} &  \textbf{26.51}* & \textbf{6.95} & \textbf{24.68}**  & 73.11 & 33.63 & 16.97 & 34.87 & 77.17 & 29.81 & \textbf{11.81} & 29.90 & 75.18 \\
        \bottomrule
        
    \end{tabular}
    \caption{Gloss to pose (G2P) model performances with different enhanced gloss as input. The original dev/test instances are split based on whether it contains tagged gloss generated by our best tagger in section \S \ref{sec:corpus_construct}. B\textsubscript{1}, B\textsubscript{4}, RG and BS refer to  BLEU-1, BLEU-4, ROUGE and BERTScore respectively. The
marks * and ** denote that the results are significant comparing to baseline with the significance level p < 0.1 and p < 0.05 respectively. Best performances are shown in bold typeface.}
    \label{tab:split_result}
\end{table*}
\subsection{Dynamic Selection Generator}

The PT baseline can generate sign poses from a single source of gloss end-to-end. However, in different scenarios, the signers may employ diverse intensification strategies to present meanings for the same gloss word (i.e. they may use a gesture with a delayed-release to represent ``heavy thunderstorm'' and later employ an end-marking to strengthen the intensity of another sign). To model this, we propose a new structure on top of the PT baseline. Given a text sequence, we mix $k$ sources of glosses with different information goals and generate signed languages that dynamically pick the source gloss. In general, we can have multiple encoders, $Encoder_{1 \cdots k}$, to encode the glosses separately and obtain the representations $src_{1 \cdots k}$. 
We utilize a single decoder to decode the output representation $k$ times from $k$ sources of encoders, each with a different encoded input representation:
\vspace{-.2cm}
\begin{equation}
    src_k = Encoder(x_{1:N}^k)
\end{equation} 
\vspace{-15pt}
\begin{equation}
    \hat{y}_{t+1}^k = Decoder(\hat{y}_{t}^k | \hat{y}^k_{1:{t-1}},src_k)
\end{equation}
We employ a multi-layer perceptron (MLP) followed by a softmax activation function to generate selection probability distributions of each source for individual frames, which we call as \textit{importance coefficients}, $IC_{t+1}$, that are conditioned on the decoded representations $\{\hat{y}_{t+1}^k\}$: 
\begin{equation}
    {IC}_{t+1} = \{\alpha_{t+1}^1, ..., \alpha_{t+1}^k\} = IC(\{\hat{y}_{t+1}^k\})
\end{equation}
 This strategy is different from \cite{Saunders_2021_ICCV} where our decoded representation $y_{t+1}^k$ aims at generating source-dependent sequences, while \cite{Saunders_2021_ICCV} applies the self-attention on the decoded sequences only. We have two variants while generating the weighted output: Dynamic and Dynamic$_{hard}$. The final dynamic output is a weighted mixture of the two candidate sequences: 
 \vspace{-.1cm}
\begin{equation}
    \hat{y}_{t+1} = \sum_{i=1}^{K} \alpha_{t+1}^k \hat{y}_{t+1}^k 
\end{equation} 
In this specific model we set the $k$ to be 2.
For the Dynamic$_{hard}$ variant of the model which picks the most plausible view at each frame as $\hat{y}_{t+1} = \hat{y}_{t+1}^k$ where $ k = \underset{i}{\arg\max}\{\alpha_{t+1}^i\}$. 

\section{Evaluations and Results}
Evaluation of sign language generation is challenging due to the lack of an automatic metric to assess the quality of generated signs. The standard practice \cite{saunders2020progressive} is to translate the poses back to the text domain and compare with ground truth text. This is called back-translation. Such automatic evaluation however, cannot accurately capture the quality of the generated signs \cite{yin-etal-2021-including}. Thus, to complement our automatic evaluation, we ask sign language experts to evaluate the generated signs. Lastly, we perform a qualitative analysis of the back translated text to i) confirm increased presence of intensity modifiers, ii) identify limitations of our models, and iii) pitfalls of existing metrics.

\subsection{Automatic Evaluation}
\paragraph{Splits and Metrics.} Prior analysis on a subset of the PHOENIX-14T's dev set unveils the imbalanced distribution of data regarding the intensity modification phenomena. Thus, results on the original data split could not faithfully evaluate the model's capability to generate intensification-specific sentences. To this end, we develop a new data split -- we collect data points which have at least one gloss labeled as either low or high intensity to construct the "with intensification" subset, and leave the remaining in a "without intensification" group. We report the BLEU-1, BLEU-4 \cite{bleu2002}, ROUGE \cite{lin-2004-rouge} on the back translated texts. We retrain the Sign Language Transformer \cite{camgoz2020sign} (SLT) to translate the sign skeletal sequences back into German texts. 
For the more fine-grained settings of \textit{intensification}-focused evaluation, we additionally report the BertScore \cite{bert-score}, an automatic metric for text generation that correlates better with human judgements, to measure the semantic similarities. We report statistical significance with
bootstrap resampling on both 90\% and 95\% confidence levels \cite{ EfroTibs93, koehn-2004-statistical}.

\begin{figure*}[!t]
    \centering
    \includegraphics[width=12cm]{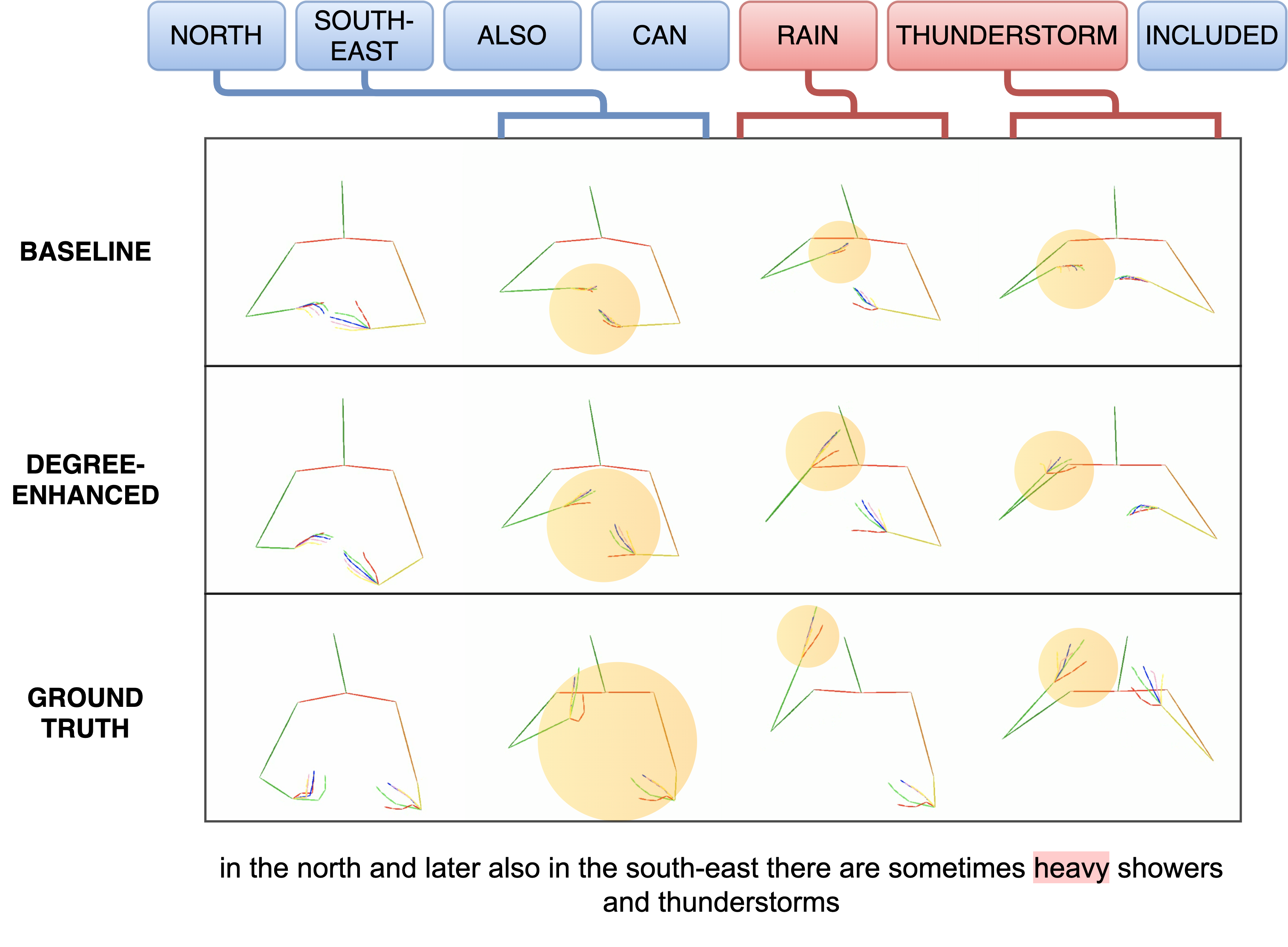}
    \caption{This figure illustrates the comparison between baseline and the intensification-enhanced model. 
    Gloss annotations are linked to their corresponding frames. Here, ground truth skeleton uses wider movements due to the "heavy" modifier, and the intensification-enhanced outputs replicate the phenomena better than baseline.}
    \label{fig:human_eval}
\end{figure*}
\paragraph{Result.}
We train a baseline PT model on the original dataset and compare it to others which are trained on the enhanced data.
We observe that, as shown in \textit{full} columns of Table \ref{tab:split_result}, the enhanced glosses improve the quality of skeleton generation on the original split of dataset. We can see that our proposed intensification enhancement techniques obtain an average of 0.6 improvement on BLEU-4 score over the dev set, with significant improvement of more than 1.6 on ROUGE. We do not observe a significant difference in the test set evaluations. Our proposed models obtain the highest ROUGE score, with negligible drop of BLEU scores comparing to models based on single source of gloss on dev set. 

Regarding the new ``with'' and ``without intensification'' splits, we first observe that there exists a considerable score difference across all three metrics between the two groups. We hypothesize that current sign language generation models are biased towards reconstructing sentences without any intensification modifiers and lack the capability to represent the intensity modification. Over the ``with intensification'' subset, most enhanced data obtain significant improvements on BLEU-1 and ROGUE score. Meanwhile, \textit{Suffixation} results in stable performance gain over the ``without intensification'' subset. This demonstrates the model's capability to distinguish between different intensified texts, such that the difference between \textit{rain} and \textit{shower} signs can be obtained while the provided glosses remain the same. The harnessing of repetitions on top of \textit{Suffixation} glosses bring in minor improvements on ``with intensification'' dev cases, and major gains are attributed to the ``without intensification'' test cases. In the end, our proposed \textit{Dynamic} model obtains the highest test set performance, where the gains are mainly attributed to the improvements over the ``with intensification'' subgroup.



    
        

\subsection{Human Evaluation}
\begin{figure}[!t]
\begin{center}
\includegraphics[scale=0.16]{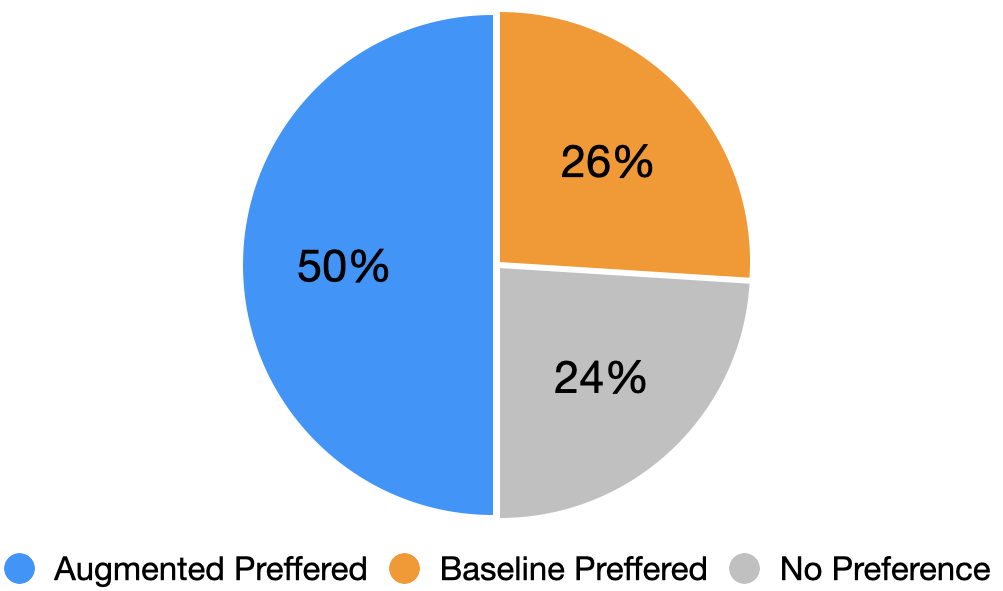}
 \caption{Human evaluation results for the generated skeletons.}
    \label{fig:evaluation}
 \end{center}
\end{figure}

\begin{table*}[]
    \centering
    \small
    \begin{tabular}{l|l|c|c|c|c}
    
    \toprule
   
    & \textbf{Examples} (Translated from German) & \textbf{B\textsubscript{1}} & \textbf{B\textsubscript{4}} & \textbf{RG} & \textbf{BS} \\
    \hline
    \multicolumn{2}{l}{\textbf{Better capture of intensity modifiers}} \\ 
    \hline
        G. Truth & The wind \textbf{usually} blows \textbf{weakly} from different directions. & - & - & - & -\\
       {Baseline} &  The wind blows \Badbox{weak to moderate} & 47.8 & 0 & 55.7 & 81.9\\
       Enhanced &  The wind \Goodboxhigher{usually} blows \Goodboxhigher{weakly} from different directions. & 100 & 100 & 100 & 100\\
        \hline
 \multicolumn{2}{l}{\textbf{Model hallucinations}} \\
    \hline
        G. Truth & The wind blows \textbf{weak to moderate} at the sea also fresh & - & - & - & - \\
       {Baseline} &  On the Alps and in the south, the wind blows \Goodbox{weak to moderate} & 50 & 0 & 46.2 & 81.7\\
       Enhanced &  The wind blows in the south weak otherwise \Goodboxhigher{weak to moderately} & 36.8 & 0 & 50.1 & 81.9\\
       & \Badbox{sometimes} fresh \Badboxhigher{to strong gusty from south to West} & & &\\
       \hline
\multicolumn{2}{l}{\textbf{Metrics failure}} \\
    \hline
        G. Truth & Tonight there are still  \textbf{a few} thunderstorms possible in the south, otherwise & - & - & - & -\\
        &rain only falls \textbf{here and there}, in places fog forms  & & & & \\
       {Baseline} &  Tonight, especially in the south and east there are rain or snow or freezing rain & 37.9 & 15.4 & 39.6 & 75.4\\
       Enhanced &  Tonight, especially in the south and east \Goodbox{here and there} \Goodbox{a few} drops or flakes & 32 & 0 & 36.9 & 75.6\\
        
     \bottomrule
    \end{tabular}
     \caption{Examples of qualitative analysis over 100 back translated texts from the videos generated by baseline and our intensification enhanced model. \textbf{Bold} texts refer to the intensity modifiers that are missing in the gloss, \Goodboxhigher{blue highlight} marks good generations and \Badboxhigher{red highlight} marks the errors.  Our model can better retain the intensity information than the baseline. Meanwhile, as shown in the third example, n-grams based metrics may fail to reward the better intensity modifier representation. }

    \label{tab:back_eval}
\end{table*}

We carry out a comparative human evaluation over 50 skeleton videos generated by both the baseline and our best performing  model 
for human annotations. For each paired video, we ask deaf sign language users to identify the video that they found to be better than the other.
They are specifically instructed to observe the following qualities and make their decisions: naturalness of the hand movements,
alignment of the hand movements (excluding finger movements) with the ground truth, representation of intensity 
by the hand movements, and overall understandability. 

As shown in Figure \ref{fig:evaluation},
outputs generated by our model trained on the enhanced glosses were preferred by signers (50\% for our model vs. 26\% for baseline). This difference is statistically different from chance as shown from a chi-squared test with p = .00017. This further suggests that a qualitative improvement using our enhancement strategies is evident. Aspects that are not fully captured by the metric-based evaluations are more clear in the human evaluations which show that incorporating intensity into the model is crucial. Enhanced glosses can generate more natural videos that depict the intensity of the signs. It should be noted that the solution to the problem at hand needs further improvement as suggested by the considerable number of "no preference" votes. 


\subsection{Backtranslation Analysis}
We hypothesize that due to enhanced glosses, there should be more intensity modifiers in the back translated text. To verify this, we compare the numbers of adjectives/adverbs in back translated text as an approximation of counting intensity modifiers. We observe that more adjectives/adverbs which appear in the original transcript are being generated in the "with intensification" partition by our model (an average of 0.79 per sentence compared to 0.75 of the baseline). As expected, we see less of a difference in the "without intensification" partition (0.87 compared to 0.86). This suggests our model is better at producing adjectives/adverbs that may act as intensity modifiers.

To better understand our model's behavior, we manually inspect 100 instances randomly drawn from the ``with intensification'' cases for a qualitative analysis. We compare the back translated texts generated by the baseline and \textit{Dynamic\textsubscript{hard}}. We evaluate the presence and correctness of modifiers instead of the overall quality of the back translated text. The key observations are: i) in 30\% of the cases, back translated text generated by our model has better representation of intensity modifiers compared to baseline, ii) in 3\% of the cases, our model hallucinates and over-generates intensity modifiers, and iii) in 23\% of the cases, at least two of the four automatic metrics did not reward \textit{Dynamic\textsubscript{hard}} for having better intensification. Table \ref{tab:back_eval} shows examples of these observations.

\section{Discussion and Conclusion}
One limitation of our study is the lack of spatial and temporal context in the automatic back-translation evaluation. The lack of a proper evaluation metric is a problem that needs to be addressed by an orchestrated effort from different fields surrounding the sign language research community. The necessity of more research in related fields is further highlighted by the fact that there are very few publicly available resources for sign language with glosses, limiting our choice and scope of datasets to the PHOENIX-14T dataset. Some corpora exists for American Sign Language such as How2sign \cite{duarte2021how2sign}, but without glosses, it renders certain sign language processing infeasible. Another limitation is the cumulative error propagation that dissipates through the intensity classifier, progressive transformer and back-translation, amplifying the total error. 
There is no dataset or method to do individual error analyses for each part of this pipeline. Thus, our error analyses were conducted in an incremental fashion as the errors in later stages of the pipeline depend on earlier errors. 

Despite these limitations, we show that the strategies of intensification, grounded in the linguistics of signed languages, contribute to the improvement of end-to-end sign language generation systems. This modeling effort is supported by our metric-based and human evaluation results. Future work needs to 
%
build on the role of intensity modifiers to further research in understating and modeling the prosody in sign language.




\section*{Acknowledgement}

This project was partly supported by the
University of Pittsburgh Momentum fund for research towards reducing language obstacles that
Deaf students face when developing scientific competencies. We also acknowledge the Center for Research Computing at the University of Pittsburgh
for providing part of the required computational
resources. The author affiliated with Gallaudet
University was partly supported by NSF Award
IIS-2118742. We would also like to thank Sarah Miller and Carly Leannah for their contributions for the human evaluation annotations.

\section*{Ethical Considerations}
Our work advocates for the need for more thoughtfulness of linguistic phenomena during the generation of sign videos. All models and analyses are built on a publicly available dataset. We acknowledge that some modules of our model depend on
pre-trained models such as word embeddings. These models are known to reproduce and
even magnify societal bias present in their original training data \cite{Li2021Robustness}.




\bibliography{custom}
\bibliographystyle{acl_natbib}
\newpage
\appendix
\label{sec:appendix}
\section{Error Analysis of Gloss Enhancement}
We manually categorized 100 errors made by our best classifier, M-BERT. The key observations are: i) 30\% of the errors are due to ambiguity that annotators may have for hard cases. E.g., "The wind blows weakly to moderately" can be annotated as either low-intensity (weakly) or no-intensity (moderately).
ii) aligning gloss tokens with text can be difficult (24\%). For example, in "partial snow or freezing rain", the classifier may consider "partial" to be aligned with rain, assigning it label of "low-intensity" (should be "no-intensity"). Further, presence of negation (e.g., "not much rain") and multiple occurrences of same word (e.g., "in the Bergland, it snows partly, on the alps it snows for a long time.") can make alignment a difficult task for the classifier, and
iii) 12\% of the errors can be attributed to noise in original PHOENIX data. E.g., the gloss representation can contain tokens that are not related to the transcript. We could not assign a specific category to 34\% of the errors.

\section{Gloss Classifier Implementation}
\paragraph{SVM Baselines} To construct the features for our text pair classification, we first concatenate the gloss token with the german text. Then we use term frequency-inverse document frequency (tf-idf) vectorizer to generate word and character n-gram vectors. These vectors are then used to train linear SVM classifiers. We use scikit-learn \footnote{\url{https://scikit-learn.org/stable/modules/generated/sklearn.svm.LinearSVC.html}} implementation with default parameters for training. The SVM models primarily serve as baselines. The SVM results are shown in Table \ref{tab:svm-classification}.
\begin{table}[!h]
\small
\begin{center}
\begin{tabular}{c|c|c|c|c}
\toprule
\bf{Model} & \bf{Features} & \bf{Prec.} & \bf{Recall} & \bf{F1}\\
\midrule
SVM & W[2-5] & 70.0 & 45.6 & 50.4\\
SVM & C[2-5] & 63.8 & 54.0 & 57.2\\
\bottomrule
\end{tabular}
\end{center}
\caption{GLOSS intensifier classification results for SVMs. W and C represent word and character.}
\label{tab:svm-classification}
\end{table}

\paragraph{FastText}
In our implementation, we use two separate embedding layers. One for the text and one for the gloss token. The embeddings for the text is averaged using pooling and then concatenated with the embedding of gloss token. This concatenated vector is then passed through a linear layer and sigmoid function to generate the predictions. We use embedding size of 100 and train for 10 epochs. We cross-entropy loss and ADAM optimizer with default learning rate. We use PyTorch \footnote{\url{https://pytorch.org/}} for our implementation. 
\paragraph{Bidirectional LSTM}
Similar to FastText, we have two separate embedding layers of size 100 for the text and the gloss token. the difference is that the output of text embedding layers are passed through a 2-layer bidirectional LSTM with hidden size of 300, dropout of 0.3. The output of the LSTM layers are then concatenated with the output  of gloss embedding layer. The concatenated output is then passed through ReLU activation function and then passed through a linear layer. Similar to FastText, we train for 10 epochs, use cross-entropy loss and ADAM optimizer with default learning rate.  PyTorch is used for implementation.

\paragraph{Fine-Tuned Transformers}
For our task. we fine-tune bert-base-multilingual (M-BERT) and german-bert-base-uncased (G-BERT) \footnote{\url{https://huggingface.co/dbmdz/bert-base-german-uncased}}. M-BERT is pretrained on Wikipedia text from 104 languages (including German). G-BERT is pretrained on Wikipedia dump, EU Bookshop corpus, Open Subtitles, CommonCrawl, ParaCrawl and News Crawl. The architecture of both models consists of 12 transformer blocks, hidden size of 768 and 12 self-attention heads. Since our task is classifying a pairs of texts, we fine-tune the models for sentence-pair classification. We use PyTorch implementation by HuggingFace \footnote{https://github.com/huggingface/transformers} for the fine-tuning. We fine-tune for 5 epochs with learning rate of 5e-05.

\paragraph{Computational resources and running time} Given our training data is small, the SVM baselines are very fast to train. They take less than 5 minutes to train. With an NVIDIA 2070 RTX GPU, the fastText and BiLSTM models take less than 10 minutes each. Fine-tuning each pre-trained BERT model with the same GPU but fewer epochs (5) take less than 10 minutes. 

\section{Dataset Statistics}
We use the publicly available benchmark, PHOENIX14T \cite{camgoz2018neural} dataset. This dataset comprises a collection of weather forecast videos in German Sign Language (DGS), segmented into sentences and accompanied by German transcripts from the news anchor and sign-gloss annotations. It contains videos of 9 different signers with 1066 different sign glosses and 2887 different German words. The video resolution is 210 by 260 pixels per frame and 30 frames per second.
The dataset is partitioned into training, validation, and test set with 7,096, 519, and 642 sentences, respectively.

\section{Transformer (Re-)Implementation}
We implemented Progressive Transformers models
for sign language generation task (\S \ref{sec:pt_baseline}) based on the code \footnote{\url{https://github.com/BenSaunders27/ProgressiveTransformersSLP}} released by \cite{saunders2020progressive}. We used the hyper-parameters from \cite{saunders2020progressive} and aimed at reproducing their reported results. To the best of our knowledge, albeit still slightly below on ROUGE-L F1 scores, our reported results on the baseline model are the nearest to the high value reported in the original paper, which does not have any checkpoint releasing. Both encoder and decoder are built with 2 layers, 4 heads and embedding size of 512. We apply Gaussian noise with a noise rate of 5, as proposed by \citet{saunders2020progressive}. All parts of the network are trained with Xavier initialisation \cite{pmlr-v9-glorot10a}, Adam optimization \cite{adam} with
default parameters and a learning rate of 1e-3. The model takes
5 hours to train on 1 NVIDIA GeForce 1080Ti GPU.
For our proposed Dynamic Selection model, to control the model size and make it a fair comparison, we halve the encoder and decoder's embedding size to 256. The Multi-Layer Percetron (MLP) model is composed of two linear layers with dimension of 256 and a ReLU activation. The model takes
8 hours to train on 1 NVIDIA GeForce 1080Ti GPU. We implemented the back-translation model on top of the original SLT code \cite{camgoz2020sign}. The transformer models are built with 1 layer, 2 head and embedding size of 128. The feature size is changed to 150, which is the sequence length of generated skeleton joints sequence. The recognition loss weight and translation loss weight are set to 5 and 1 respectively. The model takes around 1 hour for training and evaluation. All models introduced above are implemented with Pytorch \cite{paszke2019pytorch}.

\section{Parameter Comparison and Dynamic Model Experiment}
The total parameter number of each model is presented in Table \ref{tab:model_parameter}. For PT-based model, the parameter differs due to the varied size of the vocabulary sizes. Regarding the dynamic model, our early experiments show that duplicating the encoder and keeping other parameters fixed lead to worse results than the baseline model with a single encoder. This could be attributed to the limited size of our training data. We carefully tune the parameters, find that two smaller encoders could result in a stably better performance across multiple runs. 
\begin{table}[t!]
\small
    \centering
    \begin{tabular}{c|c}
    \toprule
       Model & Model Parameter \\
       \midrule
          \multicolumn{2}{l}{\textit{PT model}}\\
          \midrule
          {Baseline} & 15.3M \\
           {Suffix.} & 15.4M \\
           {Delay.-rel.} & 15.4M\\
            {End-mark.} &  15.4M\\
           {Suff.-reiter.} & 15.5M\\
     \midrule
     \multicolumn{2}{l}{\textit{Dynamic model}}\\
          \midrule
          Soft & 6.2 M\\
          Hard & 6.2 M \\

       \bottomrule
        \hline
    \end{tabular}
    \caption{Models Parameter Comparison.}
    \label{tab:model_parameter}
\end{table}

To verify the effects of mixing up two different strategies, we retrain a Dynamic\textsubscript{hard} model with duplicated suffixation enhanced data. This differs from the original model which combines suffixation and end-marking strategies. As shown in in Table \ref{tab:dynamic_result}, on the ``with intensificaiton'' split, the original \textit{Dynamic} model performs better than the one with duplicated inputs. In the ``without intensification'' split, the duplicated split gives comparable results with the baseline which is trained on the original data. 

\begin{table*}[h!]
\small
    \centering
    \setlength\tabcolsep{2.1pt}
    \renewcommand{\arraystretch}{1.1}
    \begin{tabular}{l|cccc|cccc|cccc}
    \toprule
        \multicolumn{12}{c}{\textbf{\textit{DEV SET}}} \\
        & \multicolumn{4}{c}{with intensification (248)} & \multicolumn{4}{|c}{without intensification (271)} &\multicolumn{4}{|c}{full} \\
        \midrule
         &   B\textsubscript{1} & B\textsubscript{4} & RG  & BS  &  B\textsubscript{1} & B\textsubscript{4} &  RG   & BS  &   B\textsubscript{1} &  B\textsubscript{4} & RG & BS \\

         \midrule
         
         Baseline & 25.07 & 6.24 & 22.61 & 72.20 & 35.46 & 17.98 &  36.84 & 77.46 & 29.92 & 11.90 & 30.05 & 74.95 \\
        \midrule
         Suffix. & 25.72 & 6.71 & 24.03** & 72.61 &  \textbf{37.73}** & \textbf{19.35}** & \textbf{38.92}** &  {77.88} & \textbf{31.32*} & \textbf{12.81} & {31.81}** & 75.36 \\
         \midrule
        Dynamic\textsubscript{hard} & \textbf{26.01} & 6.36 & \textbf{24.98}**  & \textbf{73.06} & 36.35 & 18.25 & 38.75** & 77.87  & 30.83 & 12.20 & \textbf{32.17}** & \textbf{75.57} \\
        
         -- two suffix. & 25.87 & \textbf{7.20} & 24.16 & 72.66 & 36.87 & 18.30 & 38.54 & \textbf{77.97} & 31.00 & 12.56 & 31.67 & 75.43 \\

        \toprule
        \multicolumn{12}{c}{\textbf{\textit{TEST SET}}} \\
        & \multicolumn{4}{|c}{with intensification (314)} & \multicolumn{4}{|c}{without intensification (328)} &\multicolumn{4}{|c}{full}\\
        \midrule
        &   B\textsubscript{1} & B\textsubscript{4} & RG  & BS  &  B\textsubscript{1} & B\textsubscript{4} &  RG   & BS  &   B\textsubscript{1} &  B\textsubscript{4} & RG & BS \\
      
        \midrule
        Baseline & 25.28 & 5.92 & 21.98 & 72.02 & 35.17 & 17.40 &  {35.97} & 76.85 & 29.86 & 11.51 & 29.13 & 74.49 \\
        \midrule
        Suffix. & 26.31 & 6.54 & {24.56}** & 73.10 & 33.70 & 17.14 & 34.60 & 76.87 &  29.73 & 11.71 & 29.69 & 75.03 \\
          \midrule
        Dynamic\textsubscript{hard} &  \textbf{26.51}* & \textbf{6.95} & \textbf{24.68}**  & \textbf{73.11} & 33.63 & 16.97 & 34.87 & 77.17 & 29.81 & {11.81} & 29.90 & 75.18 \\
      
        -- two suffix. &  26.34 & 6.82 & 24.34** & 73.10  & {34.92} & \textbf{17.46} & \textbf{36.25} & \textbf{77.49} & \textbf{30.30} & \textbf{11.94} & \textbf{30.33} & \textbf{75.35}\\
        \bottomrule
        
    \end{tabular}
    \caption{Gloss to pose (G2P) model performances on different variants of Dynamic Model. The baseline is trained using the original data. The original dev/test instances are split based on whether it contains tagged gloss generated by our best tagger in section \S \ref{sec:corpus_construct}. B\textsubscript{1}, B\textsubscript{4}, RG and BS refer to  BLEU-1, BLEU-4, ROUGE and BERTScore respectively. The
marks * and ** denote that the results are significant comparing to baseline with the significance level p < 0.1 and p < 0.05 respectively.}
    \label{tab:dynamic_result}
\end{table*}

\begin{table*}[b!]
\small
    \centering
    \setlength\tabcolsep{3.5pt}
    \begin{tabular}{c|ccccc|ccccc}
    \toprule
        & \multicolumn{5}{|c}{DEV SET} & \multicolumn{5}{|c}{TEST SET} \\
        \midrule
        Gloss Type & BLEU-1 & BLEU-2 & BLEU-3 & BLEU-4 & ROUGE &  BLEU-1 & BLEU-2 & BLEU-3 & BLEU-4 & ROUGE \\
         \midrule 
          \textit{Baseline} & 30.50 & 20.78 & 15.53 & 12.33 & 30.31 & 30.60 & 20.59 & 15.19 & 12.03 & 29.52 \\
         \midrule 
           \textit{Suffix.} &  {29.02}  & 19.88 & 14.66 & 11.66 & 29.58 & 29.30 & 19.88 & 14.66 & 11.59 & 29.28 \\
           \textit{Delay.-rel.} & 28.72 & 19.71 & 14.79 & 11.77 & 29.63 & 29.31 & 19.93 & 14.70 & 11.62 & 28.98 \\
            \textit{End-mark.} & 29.28 & 19.99 & 14.99 & 12.01 & 29.88 & 29.32 & 20.01 & 15.01 & 11.93 & 29.04 \\
           \textit{Suffix. reiter.} &  31.15  & 21.80 & 16.50 & 13.14 & 31.11 & 29.76 & 20.77 & 15.70 & 12.60 & 29.15 \\
       
       \bottomrule
        \hline
    \end{tabular}
    \caption{Translation results of the SLT model \cite{camgoz2020sign} used for back-translation. All models are trained and evaluated with ground truth hand and body skeleton joints (manual) and different choices of augmented gloss. The Baseline model is trained on the original gloss with no intensification marker.}
    \label{tab:SLT}
\end{table*}

\section{Retrained SLT model}
Given the different versions of degree enhanced dataset (\S \ref{sec:corpus_construct}, besides the baseline which is trained with the original gloss, we further retrain different versions of SLT models on the original text, skeleton joints sequence and the new gloss triples. This can serve as an estimation of the model's back translation quality given the oracle sign sequence. Table \ref{tab:SLT} shows the results.

\end{document}